\documentclass{article}
\usepackage[preprint]{icml2026}
\usepackage[utf8]{inputenc}
\usepackage{microtype}
\usepackage{graphicx}
\usepackage{float}
\usepackage{booktabs}
\usepackage{array}
\usepackage{longtable}
\usepackage{enumitem}
\usepackage{varwidth}
\usepackage{afterpage}
\newsavebox{\codebox}
\usepackage{amsmath,amssymb}
\usepackage{newtxtext,newtxmath}
\usepackage{url}
\usepackage{textcomp}
\graphicspath{{figures/}}

\icmltitlerunning{Representational drift in a code model}

\begin{document}
\twocolumn[
\icmltitle{Representational drift in a code model: \\ behaviour settles, which neurons carry it does not}
\begin{icmlauthorlist}
\icmlauthor{Piotr Wilam}{ucl}
\end{icmlauthorlist}
\icmlaffiliation{ucl}{University College London, London, United Kingdom}
\icmlcorrespondingauthor{Piotr Wilam}{piotrwilam@gmail.com}
\icmlkeywords{interpretability, language models, training dynamics}
\vskip 0.3in
]
\printAffiliationsAndNotice{}
\begin{abstract}
In neuroscience, representational drift is a change in which neurons carry a representation while behaviour stays the same; we measure its analogue in OpenCoder-1.5B during pretraining, for 169 Python constructs over two trillion tokens at a constant learning rate, with a noise unit on every change, at checkpoints as close as 1,000 training steps. For 16 of the 169, those whose grammar fixes a next token, the loss at those positions reaches its floor within the first 4\% of training, while the model's reliance on the neurons carrying them grows to the end: removing them costs 0.06, 0.25 and 0.40 nats more than removing a random set, at 4 billion, 84 billion and 2 trillion tokens. Which neurons carry a construct never settles: for all 169 the carrier set is still changing at the last checkpoint, and an abrupt settling at any date from 84 billion to 1.5 trillion tokens, or a gradual one before 500 billion, would have been detected in 99\% of injected cases. The set measured at 84 billion tokens, removed at the end, cannot be separated from a random set, so the function is carried by whichever neurons are current. Up close the change is a persisting pool and a changing sample: a neuron that leaves a construct's set returns at three times the rate of a matched neuron, and no set reassembles. The hypothesis we fixed in advance, that the carriers settle, is refuted; we discuss candidate causes and name the experiment that would separate them.
\end{abstract}

\section{The question}

A \emph{construct} is one named piece of a programming language's grammar --- the keyword \texttt{for}, the builtin \texttt{len}, a list comprehension. A \emph{neuron} is one hidden unit of a transformer's MLP block. The \emph{carriers} of a construct are the 5\% of neurons that respond most to it, measured against the same token outside its role. Once a model has learned a construct, do its carriers stay put?

On the record read here the answer is no. The model's loss at the construct's forced positions reaches its floor within the first 4\% of training; the model comes to depend more on the current carriers as training goes on; which neurons they are never settles. The separation of those three timescales, and the shape of the change up close, are the result of the paper.

In neuroscience, \emph{representational drift} is the observation that the neurons carrying a representation reorganise over days and weeks while the animal's behaviour is stable and the information stays readable from the changing population \citep{driscoll2017,rule2019,rule2020,schoonover2021}. The term is used here in that sense only: not the drift of a data distribution over time, and not the drift of a concept's meaning. For a two-layer network trained by stochastic gradient descent, an analogous drift has been derived: once a function is learned, the weights keep moving along directions that do not change the loss, faster at higher learning rates and slower for inputs seen often (\citealp{pashakhanloo2023}; from implicit regularisation, \citealp{ratzon2024}); \S5 tests both as a candidate account of what we measure.

Whether the phenomenon exists in a language model during pretraining has been seen before at coarse checkpoint spacing and without a noise unit. For subject--verb agreement, the heads that carry it in Pythia keep changing after its accuracy has settled, and the circuit shrinks from hundreds of neurons to a few dozen \citep{yao2026b}; for a set of facts in a dense 7-billion-parameter model, the 1\% of neurons that matter most change between checkpoints tens of thousands of steps apart \citep{wang2026}; for semantic roles, circuits become causally necessary early and keep refining for tens of thousands of steps, as edge weights more than as nodes \citep{aljaafari2026}. Other measurements are of the opposite: universal neurons persist across seeds at coarse spacing \citep{nandan2025,gurnee2024}, and features consolidate over pretraining \citep{bayazit2026}. \S8 shows that these and our result are one picture.

\section{How it is answered}

One record is read: OpenCoder-1.5B at 35 pretraining checkpoints spanning 4.19 billion to 2.03 trillion tokens at a constant learning rate, and 24 checkpoints of an annealing branch, which forks from it at 1.57 trillion tokens and runs to 1.67 trillion with a decaying rate. The instrument is the \emph{calibrated concept profile} of \citet{wilam2026}: a generated battery of Python prompts with every factor around the target drawn at its rate in a reference corpus, read twice (the \emph{twins}) so that every change has a noise unit, licensed on the model by a fixed set of checks, with a measured resolution across checkpoints. Appendix A gives the method in one page, and Appendix E the data behind every number; the objects it needs are these.

\begin{enumerate}[topsep=2pt,itemsep=1pt,parsep=0pt,partopsep=0pt,leftmargin=*]
\item A \emph{construct} is one piece of Python grammar or one builtin name; the battery holds 188, and 169 pass the gate that says the model's reading position is the one the prompt intends. All counts below are over the 169.
\item A \emph{neuron} is one of the 147,456 hidden units of the MLP layers (24 layers $\times$ 6,144).
\item The \emph{profile} of a construct at a checkpoint is one number per neuron: the mean standardised activation on prompts with the construct in its role, minus the same on prompts with the construct's token out of its role. The 5\% of neurons with the largest entries are the construct's \emph{carriers}; the \emph{carrier set} has 7,373 members.
\item \emph{The twins} are the battery drawn twice from one design. Their disagreement is the unit every change is measured in: within a trained checkpoint the twins' carrier sets overlap by 0.84 (the \emph{twin ceiling}; Appendix A gives the baselines); its complement, 0.16, is the \emph{twin turnover}. Both are per construct: Figure 2 subtracts the turnover at a pair's earlier checkpoint, the settling rule divides by the ceiling at its later one.
\item \emph{The six metrics} are taken for each construct at each checkpoint; one metric followed across checkpoints is a \emph{curve}. They are the turnover (item 6 below), the size of the response on the raw profile, its depth (the mean layer of its weight), the loss at the construct's marked positions (item 7 below), the sharing of carriers with other constructs, and the sparsity of the profile (the share of its squared norm in the largest 5\% of entries).
\item The \emph{turnover} at a checkpoint is the share of its carrier set that is no longer in the set at a later checkpoint, its \emph{partner}: a turnover of 0.3 means that 30\% of the 7,373 carriers have left by the partner.
\item A \emph{marked position} is a position in a prompt where the construct's grammar decides the next token: \emph{forced} (after \texttt{for x in y} the colon is obliged), \emph{expected} (what follows \texttt{for x}), \emph{bound-name} (a later use of a name the construct binds: after \texttt{for x in y:}, each \texttt{x} in the loop's body), or a scope boundary. The battery marks them for 16 of the 169 constructs, those whose grammar has such a position; the loss and the ablation below are read there, and are therefore on those 16.
\item A \emph{date} belongs to one metric: the checkpoint after which that metric's curve stops changing by more than the noise, as in the size's date or the depth's date. The turnover's date is the \emph{settling date}, the date the hypothesis of this paper is about.
\end{enumerate}

\emph{Spacing.} The record divides pretraining into three periods. \emph{Period 0}, from the start to 4.19 billion tokens, lies before the first checkpoint. \emph{Period 1}, from 4.19 to 84 billion tokens, has a checkpoint every 1,000 steps, nineteen pairs of neighbours. \emph{Period 2}, from 84 billion to 2.03 trillion tokens, has fifteen checkpoints, each with 1.25 times the tokens of the one before, so the distance between neighbours grows from 5,000 to 91,000 steps. A step is 4.19 million tokens; positions in training are given in tokens, distances between checkpoints in steps. Because the distance grows in period 2, a quantity compared between neighbours there mixes the point in training with the distance. Every rate below is therefore read at a fixed distance: for each pair of checkpoints, the change in the carrier set minus the twin turnover, within windows of the distance --- about 1,000, 5,000--10,000, 20,000--30,000 and 50,000--100,000 steps --- against the earlier checkpoint, as medians over the 169 constructs with construct-bootstrap intervals.

\emph{Resolution.} The step between a construct's profiles at neighbouring checkpoints is 8.56 twin units at the start of the record and 3.56 at the end of period 1, above one everywhere except for one construct in two pairs; so what limits the dating of a change is the spacing of the checkpoints, 1,000 training steps in period 1, not the instrument.

\emph{What we expected.} Before the record was read we assumed that, for most constructs, the carriers stop changing somewhere inside the two trillion tokens, so that a settling date exists, and we wrote nine predictions about that date and what orders it, each with its refutation condition; the analysis plan that fixed their readings was hashed. The record refutes the assumption (\S3), so most of the predictions cannot be read. They stand as written, with their verdicts, in Appendix B.

\section{The answer: three timescales}

Three things are measured for the same constructs across the same checkpoints: the loss at the marked positions (\emph{behaviour}; 16 constructs, 24 series at forced and expected positions); the rise in that loss when the current carrier set is removed, over the rise from a random set of the same size (\emph{reliance}; the same 16); and the change in the carrier set's membership between checkpoints beyond the twin turnover (\emph{membership}; 169 constructs). Figure 1 shows the three together. \S3.1--3.3 and \S3.5 follow training from its start to the end of annealing; \S3.4 gives the power of the null.

\begin{figure*}[tp]
\centering
\includegraphics[width=1\textwidth]{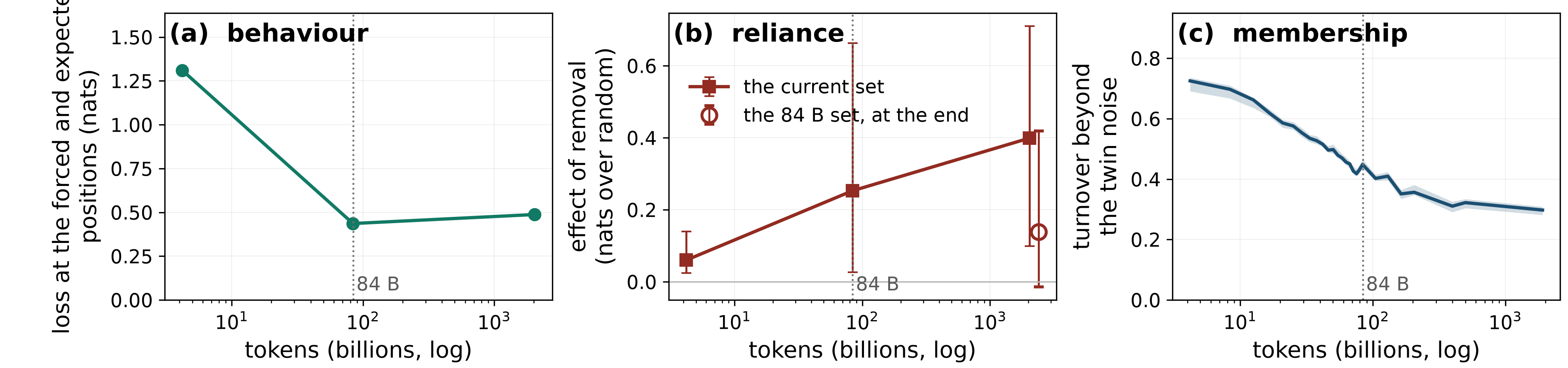}
\setcounter{figure}{0}
\caption{The three timescales. (a) Behaviour: the median loss at the forced and expected positions of the 16 constructs with marked positions, in nats, at 4.19 billion, 84 billion and 2,034 billion tokens. It falls by 84 billion tokens and is flat afterwards. (b) Reliance: the median effect of removing the current carrier set on that loss, in nats over the mean effect of removing a random set of the same size, with 95\% intervals (a percentile bootstrap of the median over the 16 constructs, 10,000 draws). The hollow marker is the set measured at 84 billion tokens, removed at the end; it cannot be separated from a random set (10 of 14 constructs above it). (c) Membership: the median change in the 5\% carrier set over 20,000--30,000 steps beyond the twin noise, over 169 constructs, with a 95\% band from the same kind of bootstrap (1,000 draws). The dotted line marks 84 billion tokens.}
\label{fig:1}
\end{figure*}

\subsection{Period 0 --- before the first checkpoint, 0 to 4.19 billion tokens, 0.2\% of training}

Everything that we expected to see forming had already formed. At 4.19 billion tokens, 167 of the 169 constructs already carry half of their profile's squared norm in 5\% of the neurons (all but \texttt{For} and \texttt{Import}), and the set is already definite: the two twins' carrier sets overlap by a median 0.84, and by more than 0.5 for all 169.

\subsection{Period 1 --- the first 84 billion tokens, 4\% of training: behaviour settles}

This is where the model finishes learning at the positions we can read. Of the 24 loss series at the forced and expected positions of the 16 constructs (one per construct, role and token; Appendix C.5), the loss falls from 4.19 to 84 billion tokens for 15, is flat for 6 and rises for 3. The four bound-name and scope positions that can be read, three bound-name and one at a scope boundary, show steep falls here too (\texttt{For}'s series alone falls from 5.6 to 2.5 nats). The carriers change fast at the start and slow down: pooled over all constructs, the fraction of carriers kept from one checkpoint to the next rises from 0.31 to 0.66 by about 71 billion tokens (Appendix D.2).

\subsection{Period 2 --- 84 billion to 2.03 trillion tokens, 96\% of training}

\textbf{Behaviour stays flat.} Over this leg the loss at the forced and expected positions is flat for 20 of 24 series, rising for 4 and falling for 0, at every threshold setting but the tightest (5\% of the remaining drop, half a noise unit), where the majority turns to rising (14 of 24). The four bound-name and scope positions are flat too, 4 of 4.

\textbf{Reliance grows.} For the 16 constructs, removing the carriers raises the loss at the marked positions by more than removing a random set of the same size does: by a median 0.06 nats at 4.19 billion tokens [0.02, 0.14], 0.25 at 84 billion [0.03, 0.66] and 0.40 at the end [0.10, 0.70], on the 20-cell yardstick (construct-bootstrap 95\% intervals; above zero for 12, 14 and 14 of the 16 constructs, sign test p = 0.038, 0.002 and 0.002; the largest effect is \texttt{Try}'s, 5.95 nats above a random set on a 0.088-nat baseline). Taken one at a time against 320 random sets, 2, 8 and 8 of the 16 constructs exceed every random set at the three references, the same after a Holm correction; the claim is the one over constructs, and Table D.1 gives each construct's rank (Appendix D.6 says why ranks and not the control's spread). The size of the effect depends on the yardstick's position cells --- under one pooled cell it is 0.73 nats at the end --- and the direction does not \citep[\S4.3]{wilam2026}. So the model relies on the neurons already at 4.19 billion tokens and more by 84 billion, and depends on them more for the rest of training, while the behaviour they serve is flat.

\textbf{Membership never settles.} For 169 of 169 constructs the turnover curve is still moving at the last checkpoint, and no construct has a settling date in pretraining under the pre-registered rule as corrected (Appendix C). The hypothesis of \S2 is refuted on this record, and the seven predictions that need a settling date cannot be read (Appendix B). At the end the change over 20,000--30,000 steps is still about twice the twin turnover (Figure 1c; \S4.1 gives the shape).

\textbf{The old carriers' function is not retained; the readout is.} The set measured at 84 billion tokens, with 14\% of its members still in the current set, raises the loss when removed at the end by 0.14 nats over the random sets [$-$0.01, 0.42], 10 of 14 constructs, p = 0.09, so it cannot be separated from a random set (4 of the 14 with an old set exceed every random set, against 8 of 16 for the current set; \texttt{For} and \texttt{If} have no old set at the last reference); against the current set it is 0.28 nats weaker [$-$0.77, $-$0.11], 11 of 14 below it. The function is carried by whichever neurons are current. What does not move is the readout: the loss at the marked positions (above), and the construct's readability by a linear probe on the activations, at a median accuracy of 0.998--0.999 at the three trained checkpoints where it was run \citep[\S4.7]{wilam2026}. This is the analogue of the neuroscience result --- stable task information from an unstable population \citep{rule2020} --- and the sense in which the title's "behaviour settles" is meant.

\subsection{The power of the null}

"No construct settles" is informative only where settling would have been found. Settlings were injected into the measured turnover series at five dates from 84 billion to 1.5 trillion tokens, as a step or as a gradual approach, with the noise of other constructs' series, and the corrected rule applied (Appendix C.3--C.4). An abrupt settling is detected in 99.1--99.4\% of 338 draws at every date; a gradual one in 99.4--100\% up to 500 billion tokens and in none at 1 and 1.5 trillion, where its approach is still under way at the end of the record; on uninjected series the rule invents none in 338. Hence "0 of 169" means: no construct settles abruptly at any date from 84 billion to 1.5 trillion tokens, and none settles gradually before about 500 billion tokens (Figure 3).

\begin{figure*}[tp]
\centering
\includegraphics[width=1\textwidth]{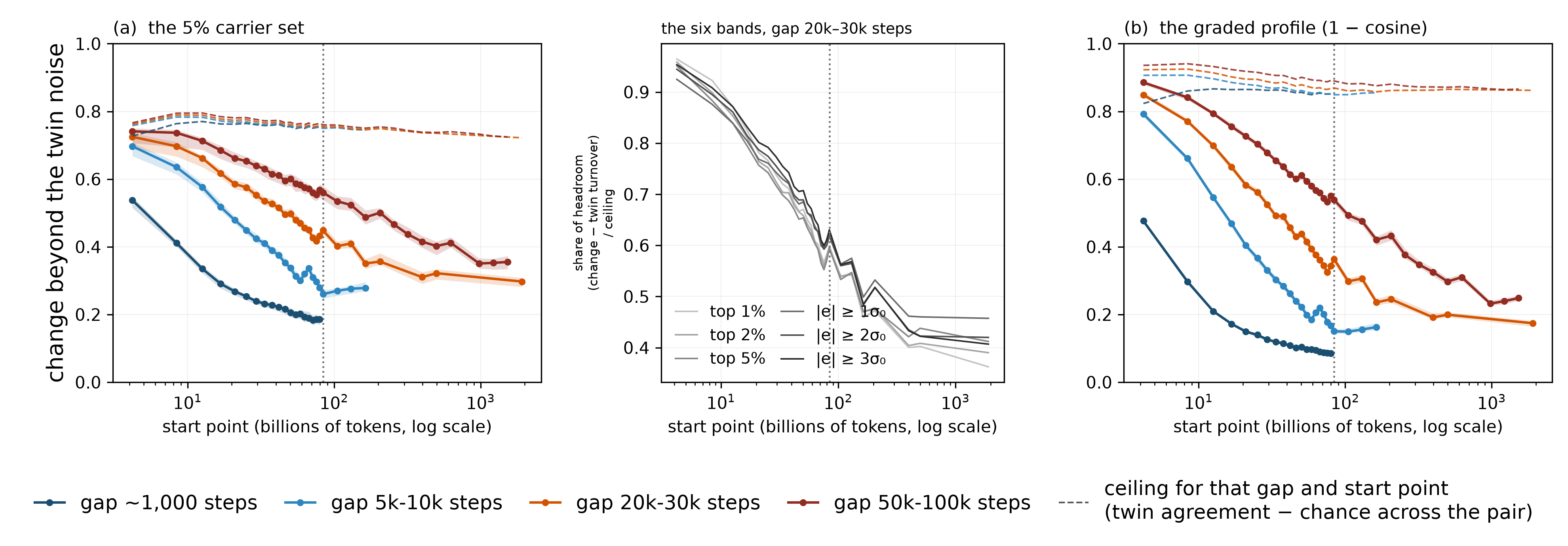}
\setcounter{figure}{1}
\caption{Change in membership beyond the twin turnover at fixed gaps: medians over 169 constructs by start point, in four windows of the gap, with construct-bootstrap 95\% bands; dashed lines are each window's ceiling (\S4.1). (a) The 5\% carrier set; below it, the six cuts at 20,000--30,000 steps as shares of their ceilings. (b) The graded profile, as 1 $-$ cosine. Dotted line: 84 billion tokens.}
\label{fig:2}
\end{figure*}

\begin{figure*}[tp]
\centering
\includegraphics[width=1\textwidth]{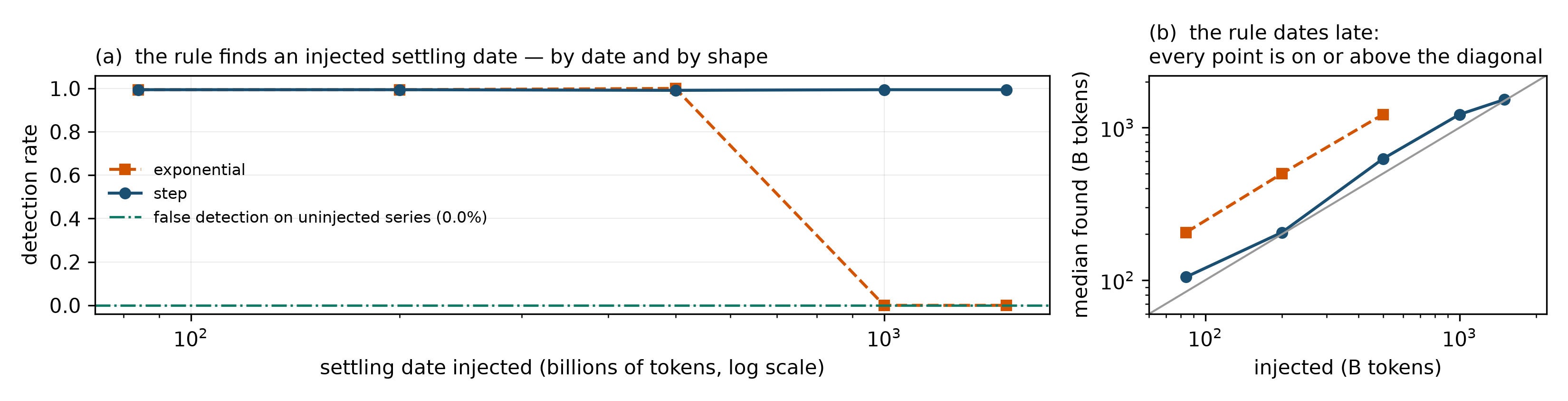}
\setcounter{figure}{2}
\caption{The power of the null. Settling curves were injected into measured turnover series at five dates, in two shapes, 338 draws a cell. (a) The share detected by the corrected settling rule; the false-detection rate on uninjected series is 0 of 338. (b) The median date found against the date injected. Every point is on or above the diagonal, so the rule dates late and any date it reports is an upper bound.}
\label{fig:3}
\end{figure*}

\subsection{The annealing branch --- forked at 1.57 trillion tokens, run to 1.67 trillion}

Here the learning rate decays, as the cooldown of a warmup--stable--decay schedule \citep{dremov2025}. Measured over equal steps of training, carrier change slows for 169 of 169 constructs. The branch is too short for dates: it spans 0.026 decades of tokens, less than a third of one pretraining checkpoint step, so a date inside it cannot be ordered against anything. The data mix also changes at the start of this branch, at the same moment as the learning rate, so the slowing cannot be attributed to the rate on this record. The loss at the forced positions partly drops on this branch: 5 of 24 series fall within it, and of the 23 with a pretraining counterpart 7 end more than one noise unit below the pretraining floor (\S5).

\section{What the change is made of}

The membership change has a shape, and the record measures six things about it. No cause is named in this section.

\subsection{At a fixed gap the change keeps slowing and never reaches the noise}

In period 1, consider two checkpoints 1,000 steps apart. The turnover between them, minus the twin turnover (0.16), is 0.54 for the first pair of the period and 0.19 for the last. In period 2 neighbours are further apart, so turnover is read at two fixed distances. For checkpoints 50,000--100,000 steps apart it falls from 0.53 at the first reading of the period to 0.35 at the last; for checkpoints 20,000--30,000 steps apart, from 0.40 to 0.30. The last reading at each distance is the latest checkpoint that still has a partner that far away. In both, the intervals at the first and last readings do not overlap. Each excess is read against its own ceiling, the most the set could change over that pair: the twin agreement minus the agreement between different constructs across the same two checkpoints. At the first start point the longest window reaches 0.97 of its ceiling, so early in training almost the whole set is replaced over 50,000--100,000 steps. The ceilings stay nearly flat over the record (0.73--0.80 for the 5\% set) while every series falls, so the slowing is not a ceiling effect. At the end the change is still about twice the twin turnover (Figure 2a).

The same holds for the whole graded profile, without any threshold: the loss of cosine similarity beyond the twin noise falls over 50,000--100,000 steps from 0.49 to 0.24, against a ceiling of 0.87--0.88 (Figure 2b), and at every one of six cuts from the top 1\% to entries above one noise unit no construct settles and the change, as a share of each cut's ceiling, is nearly the same (Figure 2, lower strip; Appendix D.2). The rate of change is a property of the profile, not of the 5\% cut. A fitted form is reported in Table D.2 only where it explains more than 80\% of the variance; the paper's reading is the fit-free one. Frequency does not order the speed of carrier change at any checkpoint: at each of the 19 checkpoints of period 1 the rank correlation between a construct's frequency and the fraction of its carriers kept has a median of $-$0.07, never significant in the predicted direction; in the late checkpoints $-$0.03. A pilot figure of +0.54 on 30 constructs does not reproduce on the 162 constructs with a frequency reading.

\subsection{Departed neurons return; the set never reassembles}

Of 18,034,712 departures of a neuron from a construct's 5\% set between neighbouring checkpoints, 57\% come back to the same set at a later checkpoint, against 17\% for neurons matched on being out of the set at the same moment (lift 3.3, interval 3.27--3.37); the median absence is 10,000 steps (quartiles 2,000--58,000), and 39\% are back at the next checkpoint (Figure 4a). This is not threshold jitter: 92.5\% of departures are falls of the neuron's entry larger than its \emph{twin difference} (Appendix A item 7), and they return at the same lift; the 7.5\% that fall by less return a little more often and sooner, as neurons at the boundary should.

The set as a whole does not return. For 167 of 169 constructs every checkpoint's nearest set is its immediate neighbour, and no two checkpoints' sets are as alike as two batteries at one checkpoint. Neurons move together only modestly, about 1.5 times what independent neurons with the same persistence would show. Together these two readings describe a pool of neurons able to carry the construct that persists, and a sample from it, the set at a checkpoint, that changes; switching among a few fixed wirings is ruled out.

\begin{figure*}[tp]
\centering
\includegraphics[width=1\textwidth]{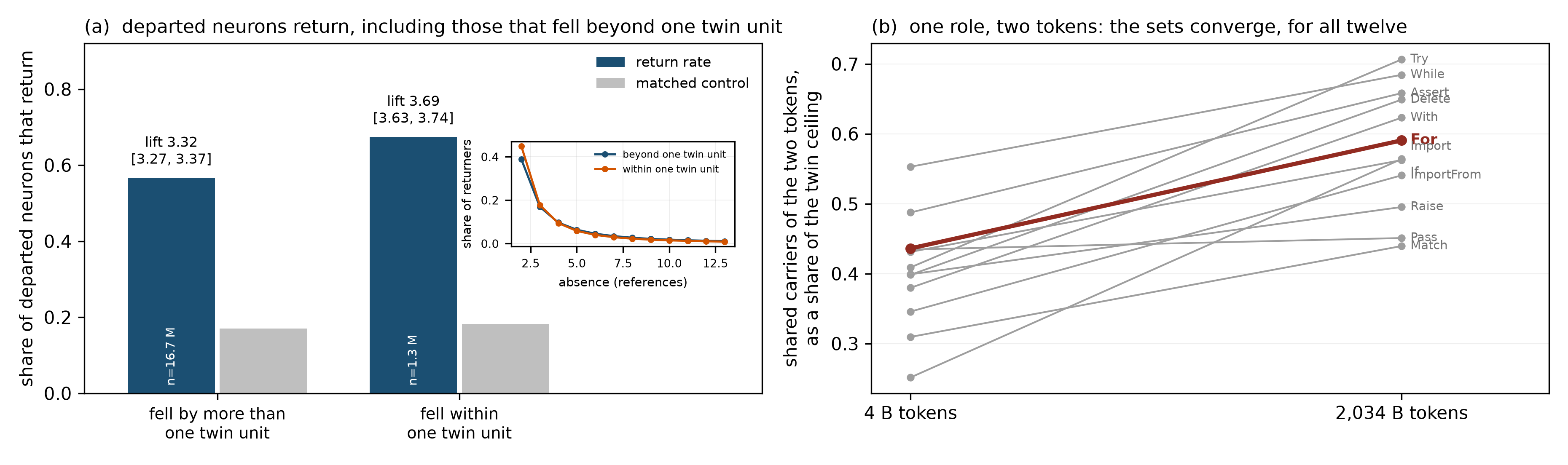}
\setcounter{figure}{3}
\caption{Two readings of the drift. (a) Returns: the share of neurons that leave a construct's 5\% set and re-enter it at a later checkpoint, against neurons matched on being out of the set at the same moment. Departures are split by whether the neuron's entry fell by more than its twin difference (16.7 million, 92.5\%) or by less (1.3 million); the lifts over the control are 3.32 [3.27, 3.37] and 3.69 [3.63, 3.74]. Inset: the distribution of absence length, in checkpoints, for the two arms. (b) Convergence of a role's two tokens: for the twelve constructs whose keyword is a different token indented and at module level, the carriers the two tokens' sets share, as a share of the twin ceiling, at 4.19 billion and 2,034 billion tokens; \texttt{For} is highlighted.}
\label{fig:4}
\end{figure*}

\subsection{The weights move as a confined walk}

Layer 12, the one layer measured, moves less between checkpoints than a free random walk would, and less as training goes on.

\emph{The weights.} The gate, up and down projections of the layer's MLP. Their movement over D steps is the size of each matrix's change, per parameter.

\emph{The free random walk.} OpenCoder trains at a constant learning rate $\eta$ = 3 $\times$ 10$^-$$^4$ after a 2,000-step warm-up \citep{huang2024}. Under an Adam-family optimiser each step moves a parameter by about $\eta$, whatever the gradient's size, so if successive steps were independent, D steps would move it $\eta$$\cdot$$\surd$D.

\emph{The result.} Over 1,000 steps the three matrices move 0.51--0.67 of $\eta$$\cdot$$\surd$D; over 20,000--30,000 steps, 0.65 early in training and 0.30 late. Successive steps partly cancel, and cancel more as training goes on: the weights move as a walk confined to a region that tightens. The carriers' rows move as random rows of the same layer do (within 1.4\%, and within 0.3\% for 15 of 24 comparisons).

\emph{The precision.} The fall from early to late holds, since every term but D is the same in both. The Frobenius norms of the three matrices grow over pretraining, by 81\%, 89\% and 83\% (used in \S5, cause 5).

\subsection{A role's two tokens converge without merging}

A statement-initial keyword is a different vocabulary entry indented and at module level (\texttt{\textvisiblespace{}for} and \texttt{for}), two embeddings as unrelated as a random pair. At the first checkpoint the two tokens' carrier sets for the \texttt{for} loop share 0.44 of what the twin ceiling allows; at the end, 0.59. Across the twelve constructs whose token differs by scope, every share is higher at the end than at the first checkpoint, running from 0.44 (\texttt{Match}) to 0.71 (\texttt{Try}) (Figure 4b). The representation moves from token towards role over training without arriving. This is the one direction of the drift the record measures cleanly; a second, towards constructs that share surrounding code, did not survive a check for the battery's composition (Appendix D.4).

\subsection{The response shrinks, and frequency orders only when the shrinking stops}

The size of the response --- on the raw, unstandardised profile --- falls over training for 148 of 169 constructs while the loss is flat, and stops falling for 90, mostly between 500 billion and 1.5 trillion tokens; between 84 billion tokens and its date it falls for all 90 and rises for none. Its depth stops changing for 59 constructs and its sparsity for 97; where all three stop, they stop at different times, spread over a factor of three in tokens, and in no fixed order. A construct the model meets ten times more often reaches the point where its response stops shrinking about 1.2 times earlier in tokens (frequency orders word acquisition the same way: \citealp{chang2022}) (coefficient on log frequency $-$0.064, interval $-$0.103 to $-$0.036); how often the token appears adds nothing once the role's frequency is in; depth, sparsity and the loss show no frequency ordering.

\subsection{Sharing between constructs falls slowly almost everywhere}

Over the record the sharing of carriers between two constructs falls by more than its noise for 11,402 of the 14,196 pairs and rises by more than its noise for 1,868, so the sets separate; when a pair's fall starts cannot be timed (Appendix D.3).

\section{What this does and does not say about why}

\emph{A discussion, not a result. The account below was written after the record was read; the readings of \S4 that it uses had each its definition fixed before it was computed, and none was pre-registered (Appendix C.1).}

\textbf{"Stopped learning", stated exactly.} At the forced and expected positions the loss on the battery falls no further than one noise unit or 10\% of its remaining drop over the leg from 84 billion tokens to the end (20 of 24 series); the model's loss on its corpus keeps falling. Whether nothing is left to learn there depends on the kind of position. At \emph{forced} positions the loss is about 0.09 nats. At \emph{expected} positions, where the grammar allows more than one continuation, the model's loss (0.55--0.59 nats after \texttt{for x}) matches the battery's own continuation entropy (0.571 nats), the loss of a reader who knows only the rates. At \emph{bound-name} positions, the later uses of a name the construct binds, the loss falls from 6.0 nats to 2.5 over training against a marginal entropy of 5.9 in the battery; the true floor there is far below the marginal, and the model is still above it. So learning has finished where a construct forces the next token and has not finished where it governs one; and the pretraining floor is the floor at the constant learning rate --- on the annealing branch 7 of the 23 forced series with a pretraining counterpart end more than one noise unit below it, with the data mix changing at the same moment.

\textbf{The question.} At the forced positions the construct's own gradient is silent from 4\% of training on. Why do the neurons that carry it keep changing, ever more slowly, for the rest of training? For a two-layer network under plain gradient descent a derivation exists: stochastic gradient descent keeps moving the weights along directions the loss does not constrain, at a rate set by the learning rate, and more slowly for inputs seen often \citep{pashakhanloo2023}. That gives two predictions for this record: frequent constructs' carriers change more slowly, and the speed of change follows the learning rate. The first is refuted: frequent constructs' carriers do not change more slowly at any checkpoint (\S4.1). The second cannot be read: the one change of learning rate coincides with a change of data, on a branch too short to date (\S3.5).

\textbf{Five candidate causes.} The discussion is limited in two ways. It treats the carrier set, not the graded profile; and it treats neurons as a group, not one at a time. Both limits are those of this paper's readings, not of the question, and they may be too restrictive: an explanation may need entries below the cut, or may differ between the large and small entries of one set, or between constructs. Within the limits, five accounts can be named, three from the literature on other systems. Each is given with what it predicts here and whether \S4 bears it out.

\begin{enumerate}[topsep=2pt,itemsep=1pt,parsep=0pt,partopsep=0pt,leftmargin=*]
\item \emph{Free diffusion.} Gradient noise moves the weights at random along directions the loss does not depend on \citep{pashakhanloo2023}. It predicts a steady speed of change at a constant rate, independent of what else is learned, lower for frequent constructs. For: the weights' steps are largely noise-like (\S4.3). Against: the change keeps slowing at a constant rate, frequency does not order it (\S4.1), and the weights fall ever further short of a free walk (\S4.3). Noise accounts for the motion being there, not for how it changes.
\item \emph{Displacement by new learning.} Other constructs claim the neurons and push the carriers out. It predicts turnover that follows how much other sets grow and how full each layer becomes. For: free neurons are used up over training, and constructs for which all other sets grow more turn over slightly more (Spearman 0.09, positive in 18 of 19 pairs). Against: turnover does not follow how fast a layer fills, and neurons in many sets turn over least (Appendix D.3). Weak support; on one run it cannot be separated from cause 1, because new learning and the learning rate change together.
\item \emph{Branching.} A carrier's meaning widens to related constructs while its response to its own weakens \citep[for vision models]{cammarata2026}. It predicts that a neuron leaving a set rises at the same step for related constructs. Against: once the battery's own composition is accounted for, departures rise for related constructs no more than for others (Appendix D.4). Unsupported here.
\item \emph{Equivalent implementations.} Many sets of neurons implement the same function and training moves among them. It predicts the readout preserved while membership changes, and departed neurons that return. For: the readout holds while the old set's function is not retained (\S3.3), departed neurons return and no set reassembles (\S4.2), the walk is confined (\S4.3). Against: nothing in this record. The best supported of the five, and the one under which the parallel with neuroscience is exact \citep{rule2020}; what the pool of \S4.2 is, as an object, this paper does not say.
\item \emph{Weight decay.} With decay at a constant rate, weights move even where the gradient is zero. The release states no optimiser or decay setting (its configuration files carry architecture and inference keys only, and the checkpoints no optimiser state); at the framework's default the decay term is a small fraction of the measured step, and the weights' norms grow (\S4.3). Not credited as a driver; not excluded as a contributor.
\end{enumerate}

\textbf{What would decide.} Causes 1 and 4 are one picture at two levels: the motion is gradient noise, confined to a region of equivalent implementations whose bounds tighten as training goes on. Cause 2 adds a weak pressure from what else is being learned. The causes need not exclude one another. Each measurement above is pooled in one of two ways, and each can hide a mixture: a median over constructs can hide constructs that move for different reasons, and a total over neurons can hide, within one set, large and small entries that move for different reasons. What confines the walk, and why the confinement tightens at a constant learning rate, this record does not say, because the amount still being learned and the learning rate change together on it. The decisive experiment is one model trained twice on one corpus, with a constant and a decaying rate, and the carriers' formation inside the record (\S7).

\section{Four mistakes in the ruler, found and fixed}

Four artefacts of measurement would each have produced settling dates or a false constancy. (1) The turnover curve's level falls with the gap at which it is read (Spearman $-$0.813 over 1,404 construct--checkpoint pairs), and the last checkpoints have closer partners because training ends, so twelve constructs appeared to settle at 1.527 trillion tokens; the rule now needs agreement between two partner spacings and two checkpoints after the crossing. (2) On a branch 6\% long nearly every turnover value is above the threshold from the first checkpoint, so 169 of 169 constructs appeared to settle at the first annealing checkpoint; the rule now needs an observed crossing with a checkpoint below it before. (3) The response's size was read standardised where the contract made raw primary: on the raw form it falls for 148 of 169 constructs where the standardised rose for 144, the frequency effect shrinks from 1.9 to 1.2, and the size no longer ranks constructs with depth and sparsity. (4) A per-step rate taken by dividing a $\times$1.25 pair's turnover by its step count assumes turnover linear in steps, which (1) says it is not; a first repair fitted the gap but forced the change to zero at zero gap, missing every short-gap point and producing a floor the data do not have; rates are now read at fixed gaps after subtracting the twin turnover, and "roughly constant" became "keeps slowing". Each is reported because the next study of the same question will meet it; Appendix D.1 gives each in full, with the dates, and the checkpoint-level hazard has a behavioural form \citep{nishida2025}.

\section{Limits, and what is planned}

The claims hold within these limits.

\begin{enumerate}[topsep=2pt,itemsep=1pt,parsep=0pt,partopsep=0pt,leftmargin=*]
\item One model, one run, one schedule.
\item The readable positions are the constructs' forced grammar, not their use, and are marked for 16 constructs, on which the behaviour and reliance claims rest.
\item The record begins at 4.19 billion tokens, so formation is not in it.
\item The annealing branch is 0.026 decades wide and changes the data mix at the same moment as the rate.
\item The null has no power past about 500 billion tokens for a gradual settling.
\item The battery is synthetic Python, shown to a model that read 607 languages \citep{huang2024}.
\item The carrier set is one cut of a graded profile at 5\%; a third of its turnover is neurons crossing that line by a small margin, though tightening the cut changes neither the rate nor the verdict, and the graded profile slows in the same way.
\item Every number is on the 20-cell yardstick of \citet{wilam2026}, whose sensitivity that paper reports: the turnover moves by 0.003--0.005 under a yardstick of token classes alone; the ablation's size moves and its direction does not.
\end{enumerate}

The claims call for the same reading on other models, on natural language and under other learning-rate schedules; the candidate causes call for experiments that change one factor at a time, the first being the two-schedule run of \S5.

\section{Related work}

\textbf{Persistence and change.} \citet{nandan2025} find that over 80\% of universal neurons persist across five GPT-2 Small seeds between 100,000 and 300,000 steps and attribute the residual loss to gradual representational drift; \citet{gurnee2024} define the neurons they measure. Neurons matched across seeds at coarse spacing and one construct's carrier set followed at 1,000-step spacing are different objects, and the results are not in conflict. \citet{wang2026} attribute 906 facts to single FFN neurons across the checkpoints of OLMo-7B: the top-1\% neurons overlap by at most 10.2\% between checkpoints tens of thousands of steps apart --- the nearest reading to ours, with a different instrument. \citet{yao2026b} follow the subject--verb agreement circuit through Pythia-1B and ten seeds of Pythia-410M: accuracy settles by step 2,000, after which the mediating heads turn over in every seed while the attributed neurons fall from 555 to a few dozen that then largely persist --- our claim at the level of heads; at the level of neurons a persistence result for a circuit of 20--40 neurons chosen by attribution, a different object from a 7,373-neuron set cut from a graded profile. \citet{aljaafari2026} track edge-attribution circuits for four semantic roles through Pythia-1B over five seeds: causally necessary by steps 32--512, top-20 node sets consolidated (Jaccard $\geq$ 0.6) by step 128 to 5,000, refinement continuing to the end as edge weights rather than nodes, read as continuous rather than as phase transitions --- a persistence result for nodes at coarse spacing, and our result at the level of edges. \citet{dahiya2026} show, on indirect-object identification in Pythia 160M--1B, that a mature circuit can conceal a transient configuration that drove behaviour earlier; transplanting the mature suppressing head into the early checkpoint recovers 35--68\% of the later gain. None of these gives a cause for the reorganisation; \S5 is an attempt at one.

\textbf{Consolidation.} \citet{bayazit2026} track crosscoder features across pretraining and report consolidation of features and of their causal importance. Here consolidation holds for the readout and fails for the membership of the carrier set; the two results are of different objects --- a learned feature against a thresholded set of neurons --- and the difference in object is the point: what consolidates is the readout, not the units.

\textbf{Stages.} Phases have been described in crosscoder features \citep{ge2026}, in the geometry of representations \citep{li2025}, in behaviour \citep{michaelov2025,chen2024}, by the local learning coefficient \citep{hoogland2025} and in the topology of concept circuits \citep{yao2026}; at the positions this record can read, every stage is over by 0.2\% of training, and what follows is continuous. Li et al.'s compression phase coincides with the stretch where the raw size falls for 148 of 169 constructs; no mechanism is claimed. \citet{howe2026} pre-registers the dating of capability emergence; Appendix B was written in the same spirit and claims no priority.

\bibliographystyle{icml2026}
\bibliography{refs}
\appendix
\section{The instrument, in enough detail to read this paper alone}

The instrument is the calibrated concept profile of \citet{wilam2026}. This appendix gives what a reader needs to judge the measurements here, in the order the method builds them.

\begin{enumerate}[topsep=2pt,itemsep=1pt,parsep=0pt,partopsep=0pt,leftmargin=*]
\item \emph{The declared stimulus distribution.} For each construct, a written list of the factors around it --- identifiers and literals, scope and nesting, companion constructs, length and surrounding code --- each drawn at its rate in a public reference corpus of 253,398 parsed Python files. OpenCoder's training data was not released, so this corpus stands in for it. A prompt is one logged draw; nothing about a prompt is chosen for a reading.
\item \emph{Three sets of prompts.} Set A holds the construct in its role (\texttt{for x in range(n):} with a body). Set B holds the construct's token outside its role --- inside a string, a comment or a name --- with companions matched to A's, a match not met for the eight compound statements (item 10). Set C is ordinary code with no target, one pool shared by all constructs. A and B hold 720 prompts each per twin, ninety from each of eight cores, for every construct.
\item \emph{The read site.} The construct's first token: \texttt{for} in a loop, \texttt{len} in a call. A construct is admitted only if its A and B prompts land on the same token piece in at least 80\% of prompts. 169 of 188 pass; they are the 169 constructs of this paper.
\item \emph{The profile.} For each of the 147,456 neurons (24 layers $\times$ 6,144 hidden units of the SwiGLU MLP, read before the down-projection), the mean activation at the read site over set A minus the mean over set B. Averaging over A removes the particular prompt; subtracting B removes the token. Taken on activations as the model produces them, this is the raw form; it gives the size of the response.
\item \emph{The yardstick.} Each activation is restated in units of the neuron's ordinary spread: z = (x $-$ $\mu$$_0$)/$\sigma$$_0$, with $\mu$$_0$ and $\sigma$$_0$ the neuron's mean and spread over set C in one of 20 cells (token class $\times$ which fifth of the prompt). The difference of item 4 taken on z is the standardised profile. An entry of 2 means the neuron responds to the role by two of its ordinary spreads more than to the token alone. Carrier sets are cut on this form.
\item \emph{The carrier set.} The 5\% of neurons with the largest standardised entries, 7,373 neurons. Two random 5\% sets overlap by 0.05.
\item \emph{The twins and their three units.} Each construct has eight structurally distinct cores, and every prompt is a core in drawn surroundings. The battery is drawn twice, on the same cores with different surroundings; the two twins' disagreement is what the surroundings alone can do. It is measured in three ways, one for each kind of object the paper compares. (a) On a profile: the \emph{twin unit}, the distance $\surd$(2(1 $-$ cos)) between the two twins' profiles, cos being their cosine \citep[\S2.4]{wilam2026}; used for the resolution (item 11). (b) On a carrier set: the \emph{twin ceiling}, the overlap of the two twins' sets, about 0.84 at any trained checkpoint against 0.28 for an untrained model, and its complement the \emph{twin turnover}, 0.16; used for every change in membership (\S2 item 4 says how each enters). (c) On one neuron: the \emph{twin difference}, the absolute difference between its two twins' entries; used for departures (\S4.2).
\item \emph{What the twins do not carry.} Because the twins share their cores, their disagreement is surface noise only; variation between cores is carried by the core bootstrap (item 9).
\item \emph{Intervals.} An interval on a number derived from one construct's profile comes from the core bootstrap: the eight cores resampled with replacement, the same resample for both twins and for every checkpoint compared. An interval on a median over constructs comes from resampling the constructs; each caption says which.
\item \emph{The licence.} Nine checks run on the pair before the battery is read \citep[\S2.5, \S4.1]{wilam2026}. Two, the nulls, are a gate --- scrambled labels and random sets must give no profile --- and both pass on OpenCoder. The other seven state the restrictions within which the profile is read: agreement between two draws, convergence as prompts are added, prompt length, re-weighting of the logged factors, the intervals' coverage, the yardstick's floor, and the companions. The companion check is not met for the eight compound statements, whose A and B prompts differ in companions; it bears on this paper only through those eight.
\item \emph{The resolution.} The step between a construct's profiles at neighbouring checkpoints, in twin units. Where it is below one, no date can fall between them. On this record the median step is 8.56 twin units between the first two checkpoints, 1,000 steps apart, and 3.56 at the end of period 1, and it is above one everywhere except for one construct in two pairs. The record therefore dates a change to the spacing of its checkpoints, 1,000 training steps in period 1; the instrument's own limit lies below it.
\end{enumerate}

\emph{The standardisation.} All carrier sets are read under the 20-cell yardstick of \citet{wilam2026} (the standardisation of each neuron by its ordinary spread at that kind of position). The standardisation is load-bearing: carrier sets are cut on the standardised profile, and Wilam (2026, \S4.3) shows what changes without it. The position axis matters less: a yardstick of token classes alone names partly different carriers \citep[\S4.3]{wilam2026}, but turnover changes under it by only 0.003--0.005. Everything below is on the standardised 20-cell profile.

\section{The pre-registration: the theory we held, the nine predictions, and their verdicts}

\emph{The hypothesis of \S2 and the predictions below were written before the OpenCoder record was read; the analysis plan that fixed their readings is hashed (Appendix C). This appendix is history: it is kept as written, with the verdicts added.}

\subsection{The picture we had}

We thought a construct's life in training goes in stages. First, some neurons begin to respond to it. Then the response concentrates on a few neurons. Then those neurons keep changing for a while --- some drop out, others come in --- while the model is still improving at the construct. Then the changing stops: the construct has settled. Our theory said why each stage happens and in what order, in four statements.

\begin{enumerate}[topsep=2pt,itemsep=1pt,parsep=0pt,partopsep=0pt,leftmargin=*]
\item Training rewards whatever lowers the next-token loss. This is a fact about the procedure, not a hypothesis.
\item The model builds the cheapest detector first. The cheapest signal for a construct is its own token, so the first carriers are token detectors.
\item Anything beyond the cheapest is built only when the loss demands it. Two uses of one token share carriers until the rarer use costs enough to force a split.
\item Once a construct works, its carriers keep drifting, pushed by the noise of training and by other things being learned, along directions that do not change the loss. The drift is slower for inputs the model meets often, and it scales with the learning rate. Settling is the drift falling below what we can measure. (This statement is taken from the published derivation of \S1; statements 2 and 3 are ours.)
\end{enumerate}

\subsection{The predictions, as written before the record was read}

\begin{enumerate}[topsep=2pt,itemsep=1pt,parsep=0pt,partopsep=0pt,leftmargin=*]
\item \emph{The premise, which we never wrote down as a hypothesis because we took it for granted:} for most constructs, the carriers stop changing somewhere inside the two trillion tokens of the record, so that a settling date exists.
\item \emph{P1 --- learning first, settling after.} By the time the carriers stop changing, the loss at the construct's positions is already flat. The rival: the loss is still falling while the carriers change, so a settling date is just a learning date.
\item \emph{P2 --- frequency sets the date.} Constructs the model meets often settle earlier. A construct made of parts (\texttt{try} with its \texttt{except}) is not delayed by its parts. The rival: a composite waits on its parts.
\item \emph{P3 --- the token, not the role.} What matters is how often the token appears anywhere, not how often the grammatical construct appears. Test: constructs sharing one token (\texttt{for} in a loop, in a list comprehension, in a generator) should settle together.
\item \emph{P4 --- the learning rate.} The speed of carrier change follows the learning rate, so every construct's turnover should bend at the same checkpoint, where the rate is turned down.
\item \emph{P5 --- frequent constructs change less.} At any one checkpoint, constructs the model meets often have slower-changing carriers.
\item \emph{P6 --- shared neurons.} Neurons in many carrier sets change more; a neuron that leaves a set goes to another set rather than to none.
\item \emph{P7 --- nothing without demand.} Once the loss is flat, the strength of the response does not keep growing.
\item \emph{P8 --- shared carriers separate only under pressure.} Two uses of one token keep sharing carriers unless the rarer use's loss falls first.
\item \emph{P9 --- sharpen, then settle.} The response concentrates on few neurons before the carriers stop changing.
\item \emph{The agreement question.} Do the six curves of one construct stop at the same time, or in a fixed order?
\end{enumerate}

We also wrote down, before reading, what count of constructs we expected to be readable (80 to 120), and that if fewer than 43 constructs had a settling date, P2 would get no verdict and the main finding would be that count.

\subsection{The verdicts}

\begin{table*}[t]
\centering
\scriptsize
\renewcommand{\arraystretch}{1.25}
\begin{tabular}{@{}>{\raggedright\arraybackslash}p{0.2000\dimexpr\textwidth-8\tabcolsep\relax}>{\raggedright\arraybackslash}p{0.2000\dimexpr\textwidth-8\tabcolsep\relax}>{\raggedright\arraybackslash}p{0.2000\dimexpr\textwidth-8\tabcolsep\relax}>{\raggedright\arraybackslash}p{0.2000\dimexpr\textwidth-8\tabcolsep\relax}>{\raggedright\arraybackslash}p{0.2000\dimexpr\textwidth-8\tabcolsep\relax}@{}}
\toprule
hypothesis & what we assumed & what we found on OpenCoder & verdict & why \\
\midrule
premise & carriers stop inside the record for most constructs & they stop for none in pretraining; annealing is too short to say & refuted & 0 settling dates of 169, on either branch \\\addlinespace[2.5pt]
P1 learning first & the loss is flat while the carriers change & flat for 20 of 24 series over 96\% of training while every carrier set changes; the carriers are relied on at both ends of that stretch, already at 4.19 B and more by 84 B & confirmed & read on the 84 B $\rightarrow$ end leg as the window, since no settling date exists \\\addlinespace[2.5pt]
P2 frequency sets the date & frequent constructs settle earlier & no settling dates to order; the size's date is ordered by frequency, weakly & could not be read & needs settling dates; the size result is a different quantity \\\addlinespace[2.5pt]
P3 the token, not the role & constructs sharing a token settle together & no settling dates; on the size's date, token exposure adds nothing to role frequency & could not be read & needs settling dates; the size reading points the other way \\\addlinespace[2.5pt]
P4 the learning rate & turnover bends where the rate drops, for every construct & carrier change slows on annealing for every construct, but the data mix changes at the same moment and the branch is a third of a step wide & could not be read & needs a learning-rate change inside a branch with nothing else changing \\\addlinespace[2.5pt]
P5 frequent constructs change less & slower carrier change for frequent constructs at every checkpoint & no relation at 19 dense checkpoints or late; sign opposite where anything & refuted & read where the ruler is clean; the pre-registered refutation condition is met \\\addlinespace[2.5pt]
P6 shared neurons & neurons in many sets change more; a neuron that leaves goes to another set rather than to none & neurons in many sets change \emph{less}; of true departures 97\% go to another set against 88\% by chance & half refuted, half confirmed & the sign of the degree effect is the opposite of the prediction \\\addlinespace[2.5pt]
P7 nothing without demand & the response's size does not grow after the loss is flat & falls for 90 of 90, rises for none & confirmed & on the raw form, the primary \\\addlinespace[2.5pt]
P8 shared carriers separate under pressure & separation preceded by the rarer use's loss falling & sharing falls for four pairs in five, slowly, as an aggregate; per pair, the start of the fall is not identified & could not be read & the prediction names the order of two events; the bootstrap shows that when a pair's sharing starts to fall cannot be timed (median interval two-thirds of the record); the size of a pair's separation is unrelated to the size of the rarer use's loss drop ($\rho$ $\approx$ $-$0.04) \\\addlinespace[2.5pt]
P9 sharpen, then settle & sharpness stops before the carriers stop & no settling dates; among size, depth and sharpness, sharpness stops first in 5 of 14 & could not be read & needs settling dates \\\addlinespace[2.5pt]
agreement question & the six curves stop in a fixed order & no construct has all six dates; size, depth and sharpness stop in no fixed order; only depth and sharpness rank constructs alike; the loss stops before all of them & mostly could not be read; the readable part says no & needs constructs with every curve dated; 0 have \\
\bottomrule
\end{tabular}
\end{table*}

\subsection{What the verdicts say about the four statements}

Statement 1 (training rewards lower loss) is a fact and was not tested.

Statement 2 (the cheapest detector first, so token detectors first) has no reading on the carriers, because P3 needs settling dates. Its one exploratory reading, on when the response's size stops, goes against it: how often the token appears adds nothing once the role's frequency is in. This is weak, and it is not the quantity the statement is about.

Statement 3 (nothing beyond the cheapest without demand) has its first support: the response's size does not keep growing once the loss is flat --- it shrinks (P7, 90 of 90). Its second test could not be read: shared carriers do separate, slowly, for most pairs, but when any one pair starts to separate cannot be timed, and how much it separates has nothing to do with how much the rarer use's loss fell; the separation looks like a general spreading of the sets, and the demand account has nothing in its favour on this record.

Statement 4 (drift) is the one the record speaks to most, in two opposite ways. It is the only statement that explains the premise failure (\S5). Its two testable parts both fail to earn support on the OpenCoder record alone: the learning-rate part cannot be read, because the one rate change coincides with a data-mix change on a branch too short to date; and the frequency part is refuted on the carriers, since frequent constructs' carriers do not change more slowly at any checkpoint. A second schedule is needed for the learning-rate prediction (\S5).

\section{The record and the rules}

\subsection{What was fixed before reading, and what was corrected after}

Before the OpenCoder record was read, three things were written down and hashed: the hypothesis and nine predictions of Appendix B; the analysis plan that fixed their readings (sha256 \texttt{33b0721b\ldots{}9f81fd18}, approved 9 Oct 2026); and the design contract (\texttt{98bbc437\ldots{}3d43bd053}, 1 Oct 2026). The settling rule in that plan is C.2. It was corrected after reading, in four steps, each recorded with its date (D.1); the corrected rule is C.3. The readings of \S4.2--4.3 and \S5 --- the return rate, the recurrence of sets, the weight norms and steps --- were designed after the record had been read, with each definition written to its artefact before it was computed. The power analysis of \S3.4 was run on the corrected rule, after it was fixed. So the paper's pre-registered part is the hypothesis and the predictions; the rule that produced "0 of 169" is the corrected one, and its power was measured on the series it was applied to.

\subsection{The settling rule as pre-registered}

A construct is settled at a checkpoint if the kept fraction of its 5\% carrier set, measured against a partner checkpoint, crosses a threshold $\theta$ and stays above it to the end of the series. $\theta$ = 0.70, the contract's fallback inherited from the pilot, and the reading was repeated over a grid from 0.50 to 0.95 in steps of 0.01, so that $\theta$ is a sensitivity and not a hinge. The partner is one checkpoint at a token factor inside the band 1.25--2.00.

\subsection{What went wrong, and the rule as corrected}

The rule of C.2 was applied in two forms, which differ in how the partner checkpoint is chosen.

\begin{enumerate}[topsep=2pt,itemsep=1pt,parsep=0pt,partopsep=0pt,leftmargin=*]
\item \emph{The ratio form.} The partner is the checkpoint whose token count is between 1.25 and 2 times the checkpoint's own.
\item \emph{The fixed-step form.} The partner is the checkpoint a fixed number of training steps later.
\item \emph{What went wrong in the ratio form.} Twelve constructs dated at one and the same checkpoint, 1.527 trillion tokens. The checkpoint itself is healthy: its twin ceiling, 0.747, sits between its neighbours' 0.744--0.749. The cause is the partner. The kept fraction falls with the distance to the partner (Spearman $-$0.813 over 1,404 checkpoint--partner pairs over all constructs; median kept fraction 0.73 at a factor of 1.2--1.35 and 0.61 at 1.65--2.1). Near the end of training only the nearest partner is inside the band, because the farther ones lie past the end, so the last points are read at a shorter distance than their neighbours and look settled. One construct reads 0.611 at a factor of 1.56 and 0.717 at 1.25 at the same checkpoint.
\item \emph{What went wrong in the fixed-step form.} The same defect at a different place: its 13 dates all fall at the checkpoint whose fixed partner is the shortest gap in the tail, 29,865 steps, likewise the last point of its series.
\item \emph{A second defect in both.} "Stays above $\theta$ to the end" is empty at the last point of a series, where there is no later point to check.
\item \emph{The corrected rule.} A date needs all of: (i) an observed crossing of $\theta$, not a series already above it; (ii) at least one checkpoint below $\theta$ before the crossing; (iii) at least two checkpoints after it; (iv) agreement between the two forms; (v) in the ratio form, the kept fraction above $\theta$ at every available distance at every checkpoint from the crossing to the end, the band being split into four (1.25--1.35, 1.35--1.5, 1.5--1.65, 1.65--2.0) because one threshold cannot serve both ends.
\item \emph{Censoring.} A date in the last two checkpoints cannot satisfy (iii) and is recorded as \emph{censored}: no verdict, rather than a date.
\end{enumerate}

\subsection{What the corrected rule yields, and why the zero is informative}

Over 169 constructs the ratio form dates none (159 censored, 10 transient crossings) and the fixed form dates 13, all at the checkpoint named above; the two agree on none, so the rule dates 0 of 169, and 0 of 169 in each of the six cuts of D.2. The zero is not a rule that cannot date: the same rule, applied to the same series with a settling injected, finds an abrupt settling at 99.1--99.4\% at every date and a gradual one at 99.4--100\% up to 500 billion tokens, and invents none in 338 uninjected series (\S3.4, Figure 3). What it says is that the one form that dates 13 constructs does so at the series' last usable point, where the test cannot see the crossing hold. \texttt{DATES\_TABLE.json} gives both forms per construct with the reason no date exists.

\subsection{The loss positions}

Curve 4's series are keyed by construct, position class, role and token piece. The obliged class is the forced and expected positions; three forced positions, the header colons of \texttt{For}, \texttt{AsyncFor} and \texttt{If}, realise several token pieces in the record (the colon fuses with what follows) and are labelled \emph{merged}. The obliged class is read first: of the 38 obliged roles, those whose construct is in the 169-construct set and whose series covers all 35 pretraining checkpoints with consistent cores are 24, and these are the 24 series of \S3.2--3.3; 21 of them realise one token piece, the strict subset on which the zero floor is exact. A leg is flat if its change is within the larger of 10\% of the remaining drop at the first checkpoint and one noise unit; nine settings (5, 10, 20\% $\times$ 0.5, 1, 2 units) were run. The dependent classes, governed (the bound-name positions of \S2) and scope, are read only where the role gap exceeds one noise unit over the whole curve (rule C0.14 of the plan); a series that fails is recorded as not counting, not as a result. In the 169-construct set there are 8 governed and 3 scope series, of which 3 and 1 count: the four dependent positions of \S3.2--3.3.

\subsection{Table C.1}

The six curves' dates for all 169 constructs under the corrected rule, with both curve-1 forms and the reason no settling date exists, is \texttt{validation/paper2/DATES\_TABLE.md}; curve 5 has no per-construct date, being defined on the 14,196 construct pairs. The deviation log, with the four rule changes and every other departure from the plan, accompanies the released record.

\section{Supporting readings}

\emph{Moved from the body to keep it short; nothing here is needed to follow the argument.}

\subsection{The four mistakes in the ruler}

\begin{enumerate}[topsep=2pt,itemsep=1pt,parsep=0pt,partopsep=0pt,leftmargin=*]
\item \emph{The tail spacing.} The turnover curve's level depends on how far apart in tokens a checkpoint and its comparison partner are (Spearman $-$0.813 over the 1,404 construct--checkpoint pairs with a partner inside the band, over all 169 constructs). The last checkpoints have closer partners because training ends, so twelve constructs appeared to settle at 1.527 trillion tokens. They had not. Rule changed: a date needs agreement between two partner spacings and at least two checkpoints after the crossing. No date survived.
\item \emph{The annealing threshold.} On a branch only 6\% long, nearly every turnover value is above the settling threshold from the first checkpoint, so 169 of 169 constructs appeared to settle at the first annealing checkpoint. Rule changed: a date needs an observed crossing of the threshold, with at least one checkpoint below it before. No date survived.
\item \emph{The two forms of size.} The response's size can be measured raw or standardised. Our contract made raw primary, because the pilot had shown that the standardised size grows only because its baseline spread shrinks; the code read standardised. Re-read on the raw form: the size falls over training for 148 of 169 constructs where the standardised rose for 144; the frequency effect shrinks from a factor of 1.9 to 1.2; the \texttt{for} family's ordering disappears (1 of 6 dated, so there is no ordering to test); the size no longer ranks constructs with depth and sparsity.
\item \emph{The per-step figure from divided pairs.} An earlier draft said the change per 1,000 steps was "roughly constant across pretraining". True 1,000-step pairs exist only to 84 billion tokens; the later figure came from dividing a $\times$1.25 pair's turnover by its step count, which assumes turnover linear in steps, and artefact 1 says it is not. A first repair fitted every pair with the gap as an explicit variable, but forced the change to zero at zero gap; since two batteries at one checkpoint already differ by the twin turnover, it missed every short-gap point and produced a floor the data do not have. Rule changed: rates are read at fixed gaps after subtracting the battery's own noise, not fitted (\S4.1). The constant became a continued slowing.
\end{enumerate}

The same hazard appears in behaviour: \citet{nishida2025} report checkpoint-to-checkpoint volatility in benchmark scores large enough that averaging neighbouring checkpoints is recommended; an adjacent-checkpoint measure of any kind needs a stated noise level. A fourth check: choosing one "verdict" layer per construct is not robust --- excluding the last layer moves the chosen layer for 41\% of constructs --- so no date is ever reported from a single layer; all comparisons are made in the product space of all 147,456 neurons.

\subsection{The carrier set's change does not depend on the cut}

Pooled over all constructs, the fraction of carriers kept between neighbouring checkpoints falls from 0.66 between the checkpoints at 67 and 71 billion tokens to 0.49 between those at 1.5 and 1.9 trillion, but that figure mixes the point in training with the growing gap between checkpoints (D.1, artefact 4); the fixed-gap reading of \S4.1 does not. The change also does not depend on where the cut is made. Carrier sets cut at the top 1\%, 2\% and 5\% of neurons, and at entries above 1, 2 and 3 noise units --- sets of 744 to 14,424 neurons, from 3 to 22 times their chance overlap --- none settles for any construct under the corrected rule (0 of 169 in all six bands), and normalised by each cut's own ceiling the change is nearly the same at every cut: over 20,000--30,000 steps from the first checkpoint it is 0.925--0.964 of the ceiling across the six cuts (the broadest cut, entries above one noise unit, lowest at 0.925 [0.917, 0.937]; the top 1\% highest at 0.964 [0.952, 0.977]), whose ceilings run from 0.70 to 0.82, and the six fall together over the record (Figure 2, lower strip). The rate of change is a property of the profile, not of the cut. One small dependence on entry size exists: over long gaps ($\times$1.5), large entries keep their relative order slightly better than small ones (rank stability 0.85 against 0.82).

\textbf{Table D.2 --- the fitted per-step rates, where the fit holds.} The secondary reading of \texttt{RATE\_READINGS.json}: turnover(t, D) = c(t) + A(t)(1 - exp(-r(t) D)), with \texttt{c(t)} fixed at the measured twin turnover and fitted on the original scale by nonlinear least squares. A rate is reported \textbf{only where R$^2$ > 0.8}; a dash is a start point whose fit did not clear the floor, and its R$^2$ is shown so the reader can see how far it fell short. The primary reading of the paper is fit-free (Figure 2); this table exists so the fitted form can be checked rather than trusted.

\begin{table*}[t]
\centering
\scriptsize
\renewcommand{\arraystretch}{1.25}
\begin{tabular}{@{}>{\raggedright\arraybackslash}p{0.1667\dimexpr\textwidth-10\tabcolsep\relax}>{\raggedright\arraybackslash}p{0.1667\dimexpr\textwidth-10\tabcolsep\relax}>{\raggedright\arraybackslash}p{0.1667\dimexpr\textwidth-10\tabcolsep\relax}>{\raggedright\arraybackslash}p{0.1667\dimexpr\textwidth-10\tabcolsep\relax}>{\raggedright\arraybackslash}p{0.1667\dimexpr\textwidth-10\tabcolsep\relax}>{\raggedright\arraybackslash}p{0.1667\dimexpr\textwidth-10\tabcolsep\relax}@{}}
\toprule
start (B tokens) & iteration & 5\% set: r per 1,000 steps & 5\% set: R$^2$ & cosine: r per 1,000 steps & cosine: R$^2$ \\
\midrule
4 & 1,000 & --- & 0.668 & --- & 0.752 \\\addlinespace[2.5pt]
8 & 2,000 & --- & 0.656 & --- & 0.786 \\\addlinespace[2.5pt]
13 & 3,000 & --- & 0.698 & 0.1180 & 0.856 \\\addlinespace[2.5pt]
17 & 4,000 & --- & 0.733 & 0.0856 & 0.892 \\\addlinespace[2.5pt]
21 & 5,000 & --- & 0.769 & 0.0686 & 0.912 \\\addlinespace[2.5pt]
25 & 6,000 & --- & 0.793 & 0.0582 & 0.921 \\\addlinespace[2.5pt]
29 & 7,000 & 0.1022 & 0.812 & 0.0505 & 0.926 \\\addlinespace[2.5pt]
34 & 8,000 & 0.0935 & 0.822 & 0.0442 & 0.929 \\\addlinespace[2.5pt]
38 & 9,000 & 0.0895 & 0.823 & 0.0404 & 0.930 \\\addlinespace[2.5pt]
42 & 10,000 & 0.0873 & 0.828 & 0.0353 & 0.933 \\\addlinespace[2.5pt]
46 & 11,000 & 0.0821 & 0.834 & 0.0310 & 0.937 \\\addlinespace[2.5pt]
50 & 12,000 & 0.0807 & 0.830 & 0.0283 & 0.936 \\\addlinespace[2.5pt]
55 & 13,000 & 0.0757 & 0.835 & 0.0250 & 0.940 \\\addlinespace[2.5pt]
59 & 14,000 & 0.0750 & 0.821 & 0.0226 & 0.942 \\\addlinespace[2.5pt]
63 & 15,000 & 0.0701 & 0.814 & 0.0206 & 0.943 \\\addlinespace[2.5pt]
67 & 16,000 & --- & 0.799 & 0.0194 & 0.946 \\\addlinespace[2.5pt]
71 & 17,000 & 0.0498 & 0.801 & 0.0179 & 0.950 \\\addlinespace[2.5pt]
75 & 18,000 & --- & 0.798 & 0.0168 & 0.952 \\\addlinespace[2.5pt]
80 & 19,000 & 0.0379 & 0.808 & 0.0161 & 0.950 \\\addlinespace[2.5pt]
84 & 20,000 & 0.0352 & 0.813 & 0.0156 & 0.945 \\\addlinespace[2.5pt]
105 & 25,000 & 0.0301 & 0.813 & 0.0132 & 0.951 \\\addlinespace[2.5pt]
130 & 31,000 & 0.0255 & 0.823 & 0.0116 & 0.958 \\\addlinespace[2.5pt]
164 & 39,000 & 0.0220 & 0.821 & 0.0101 & 0.960 \\\addlinespace[2.5pt]
206 & 49,000 & 0.0201 & 0.822 & 0.0091 & 0.960 \\\addlinespace[2.5pt]
256 & 61,000 & 0.0177 & 0.824 & 0.0082 & 0.959 \\\addlinespace[2.5pt]
319 & 76,000 & 0.0153 & 0.829 & 0.0073 & 0.962 \\\addlinespace[2.5pt]
398 & 95,000 & 0.0136 & 0.842 & 0.0068 & 0.964 \\\addlinespace[2.5pt]
499 & 119,000 & 0.0121 & 0.856 & 0.0062 & 0.966 \\\addlinespace[2.5pt]
625 & 149,000 & 0.0117 & 0.859 & 0.0058 & 0.960 \\
\bottomrule
\end{tabular}
\end{table*}

\textbf{Rates reported: 21 start points for the 5\% set and 27 for the cosine, of 29 fitted. Median R$^2$: 0.821 and 0.943.}

\subsection{Sharing between constructs, and the neurons' side}

\textbf{Shared carriers do separate, slowly and almost everywhere.} A natural expectation is that two uses of one token keep sharing carriers until the rarer use's loss falls. Two constructs share a median 0.130 of their carriers against 0.050 by chance (over the 169 constructs' 14,196 pairs; 0.117 over all 188), and over the record the sharing falls beyond its noise for 11,402 of the 14,196 pairs (median fall 0.034, eight times the noise) and rises for 1,868. Under a core bootstrap of 1,000 draws on 120 pairs, 115 fall in at least two draws and the median pair falls in 88\% of them. Whether the fall follows the rarer construct's loss cannot be read: when any one pair's sharing starts to fall cannot be timed --- the interval on the beginning spans a median 30 of 35 checkpoints --- and how much it falls is unrelated to how much the rarer construct's loss drops ($\rho$ $\approx$ $-$0.04).

\textbf{Seen from the neurons' side, the carrier sets are a concentrated, committed population.} At 4.19 billion tokens 23\% of neurons belong to no carrier set (independent sets would give 0.02\%) and by the end 11\%; three neurons end in every set. Of every neuron that leaves a set, a third is still within the boundary band of the same set, between the 5\% and 10\% cut (half, when checkpoints are 1,000 steps apart); measured against noise instead, 92.5\% of departures are falls larger than the twin difference (\S4.2); nearly all the rest move into another construct's set (97\%, against 88\% by chance), and 2\% leave every set. Neurons in many sets churn least (rank correlation between a neuron's number of sets and its churn, $-$0.175), the opposite of what competition for neurons would predict.

\subsection{An exploratory reading that did not survive: where departing neurons go}

For each neuron leaving a construct's carrier set between neighbouring checkpoints, we asked whether another construct's entry on that neuron rose at the same step by more than a matched chance level, and whether the rising construct was related to the one left; relatedness, chance and the reading rule were fixed before reading (Appendix C). On three pairs of period 1, departures rose towards constructs that share a token at a lift of 1.41, 1.56 and 1.77 over chance, and towards the same grammatical family at chance. The token lift had a candidate cause in the battery itself: a token present in a construct's A prompts and absent from its B prompts --- \texttt{range} in \texttt{for}'s cores --- puts the neurons that carry it into the construct's set. Restricted to pairs where no shared token sits on one side only, the lift is 0.65, 0.79 and 0.86, below chance. The direction was the battery's composition, not the model's; we report it so that the next reading of departures avoids the same trap. On this battery a token-based relatedness cannot be separated from shared stimuli, because every one of the 49 token-related pairs shares its token in the cores.

\subsection{The battery on a second model}

The one second model already read is Pythia-1.4B, under a different tokenizer: 166 of 188 constructs pass the gate. The gate is of similar strength and fails on a different family: the five comprehension and conditional-expression constructs are lost, because the tokenizer splits \texttt{for} mid-expression, and \texttt{FunctionDef} and \texttt{AsyncFunctionDef} are gained. A second model keeps the twelve scope-varying constructs of \S4.4 but may not keep the comprehension pairs.

\subsection{The old set's effect, per construct}

Table D.1 (\texttt{validation/paper2/TABLE\_D1\_ablation.md}) gives, for the 16 constructs whose positions can be read, the effect of removing the current set at the three references and of removing the set measured at 84 billion tokens at the end, in nats over the random sets' mean, with each set's rank among 320 random sets of the same size and the exact one-sided p (a clean sweep is p = 1/321). The spread of the random sets, used as a unit in an earlier draft, is long-tailed at 84 billion tokens and at the end and does not settle with more draws, which is why the body uses nats and the table uses ranks.

\section{Data and code}

The data this paper reads are the dataset released with \citet{wilam2026}: the battery, the licence tables, the carrier sets, the record slice and the artefacts both papers read, at tag v1 of https://huggingface.co/datasets/piotrwilam/AtlasOpenCoder, DOI 10.57967/hf/11472. This paper's scripts are at https://github.com/piotrwilam/AtlasRepresentationalDrift. The dataset's README links a manifest that lists every number in this paper with the file and field it is read from and the script that rebuilds it; the analysis plan (sha256 \texttt{33b0721b\ldots{}9f81fd18}), the design contract (\texttt{98bbc437\ldots{}3d43bd053}) and the deviation log that records the four rule changes of \S6 are in its \texttt{preregistration/} directory.

\end{document}